\documentclass[conference]{IEEEtran}
\IEEEoverridecommandlockouts
\usepackage{subfig}
\usepackage{placeins}
\usepackage{cite}
\usepackage{multirow}
\usepackage{enumitem}
\usepackage{hyperref}
\usepackage[font=itshape]{quoting}
\usepackage{wrapfig}
\usepackage{float}
\usepackage{subfig}
\usepackage{multicol}
\usepackage{blindtext}
\usepackage{amsmath,amssymb,amsfonts}
\usepackage{siunitx}
\usepackage{algorithmic}
\usepackage{listings}
\usepackage{graphicx}
\usepackage{textcomp}
\usepackage{systeme}
\usepackage{hyperref}
\usepackage{xcolor}
\usepackage[LGR,T1]{fontenc}
\usepackage{mathtools}

\DeclarePairedDelimiter\floor{\lfloor}{\rfloor}

\begin{document}
\title{A Novel Time Series-to-Image Encoding Approach for Weather Phenomena Classification}
\author{\IEEEauthorblockN{Christian Giannetti}
\textit{Sapienza University of Rome}\\
\textit{giannetti.1904342@studenti.uniroma1.it}
}

\maketitle
\begin{abstract}
Rainfall estimation through the analysis of its impact on electromagnetic waves has sparked increasing interest in the research community. Recent studies have delved into its effects on cellular network performance, demonstrating the potential to forecast rainfall levels based on electromagnetic wave attenuation during precipitations. This paper aims to solve the problem of identifying the nature of specific weather phenomena from the received signal level (RSL) in 4G/LTE mobile terminals. Specifically, utilizing time-series data representing RSL, we propose a novel approach to encode time series as images and model the task as an image classification problem, which we finally address using convolutional neural networks (CNNs). The main benefit of the abovementioned procedure is the opportunity to utilize various data augmentation techniques simultaneously. This encompasses applying traditional approaches, such as moving averages, to the time series and enhancing the generated images. We have investigated various image data augmentation methods to identify the most effective combination for this scenario. In the upcoming sections, we will introduce the task of rainfall estimation and conduct a comprehensive analysis of the dataset used. Subsequently, we will formally propose a new approach for converting time series into images. To conclude, the paper's final section will present and discuss the experiments conducted, providing the reader with a brief yet comprehensive overview of the results.
\end{abstract}
\section{Introduction}
\subsection{Rainfall estimation problem}
The monitoring of rainfall holds great significance in various fields such as agriculture, runoff, flood prediction models, and the design of water-powered structures. For example, in agriculture, farmers rely on weather conditions, particularly rainfall patterns, to achieve abundant and high-quality yields. By predicting rainfall conditions, farmers can determine the most optimal times for planting and harvesting. Unpredictable variations in rainfall can disrupt cultivation schedules, leading to crop failure and significant economic losses. Rainfall estimation refers to the task of predicting the future state of the earth's atmosphere in specific locations. The primary techniques for rainfall estimation include the use of rain gauges, weather radars, and satellites. In the following paragraphs, we will briefly explore the advantages and limitations of these methods.
\hfill \break \indent
Rain gauges are commonly used to measure rainfall with high accuracy and temporal resolution. However, they only provide data for the location where they are installed. Therefore, it is necessary to use neighboring rain gauge data to create a comprehensive view of rainfall patterns. A limitation of this approach is that as the number of gauges in an area decreases, the error from interpolation increases. Additionally, a region's geographical and morphological characteristics significantly impact the accuracy of the measurements. As a result, the density of rain gauges in a region is crucial for effectively monitoring spatial rainfall distribution, as supported by various research studies. For example, Mishra et al. \cite{mishra2013effect} demonstrated that reducing the number of gauges from seven to one increased the absolute error from 15\% to 64\%.
Furthermore, the combination of low rain gauge densities and high precipitation intensities profoundly degraded the efficiency of the collected rain data, as pointed out by Girons et al. \cite{girons2015location}. An undoubtedly straightforward but costly solution could be increasing the number of rain gauges employed. However, Griffiths et al. \cite{griffiths2009spatial} have shown that even a large density of rain gauges over an area could lead to unprecedented errors during extreme precipitation phenomena.
\hfill \break \indent
Weather radars are widely used for measuring atmospheric precipitations with high temporal resolution. They work by using microwave pulses to measure the reflected signals from raindrops. By calculating the round-trip time of the radar microwave pulses, the distance to the rain can be determined. This method allows for monitoring larger areas compared to other techniques and understanding precipitation distribution over a wide area. However, it is essential to note that this type of solution can be expensive and is often affected by various sources of errors, as Jurczyk et al. \cite{jurczyk2020quality} noted.
\hfill \break \indent
Satellites are another widely used method for estimating rainfall, offering improved spatial and temporal resolution. However, they may be less accurate in estimating rainfall intensity. Therefore, this technique is commonly used to fill in radar coverage gaps and support the post-processing of radar data. As noted by Levizzani et al. \cite{levizzani2001precipitation}, the main limitation of this approach is the indirect nature of the retrieval, which links cloud properties with ground-level precipitations.
\hfill \break \indent 
The techniques for monitoring rainfall based on weather radars or satellites make use of advanced precipitation estimation algorithms (e.g., Russel et al. \cite{russell2010radar}, Thies et al. \cite{thies2011satellite}), which have shown significant improvements in performance in recent years. However, Jeon et al. \cite{jeon2019new} and Reddy et al. \cite{reddy2019evaluation} have noted that these methods are associated with multiple uncertainties, mainly when performing estimations at high resolutions. Conversely, the concept of using power loss estimation over a wireless microwave link (MWL) for rainfall estimation has been around for about 15 years, as evidenced by several research studies in this area (e.g., Marzano et al. \cite{marzano2002ground}, Leijnse et al. \cite{Leijnse_overeem2016two}, and Zinevich et al. \cite{zinevich2008estimation}). \indent 
The research community has recently developed several theoretical models that link signal attenuation to rainfall intensity. This attenuation parameter is crucial for frequencies above 10 GHz. To the best of the author's knowledge, only a few research studies have explored Earth-to-Earth rainfall microwave attenuation measurements at frequencies below 6 GHz. Most studies focus on measuring the received signal strength indicator (RSSI) on mobile terminals in the global system for mobile communications (GSM) band and estimating RSSI for wi-fi links operating at 2.4 and 5 GHz frequency bands. For example, using a probabilistic neural network, Beritelli et al. \cite{beritelli2018rainfall} have introduced a rain estimation system based on four rainfall intensity classes in 4G/LTE mobile terminals. The proposed classification method relies on three received signal level (RSL) local features of the 4G/LTE network. During meteorological phenomena, Fang et al. \cite{fang2015impact} have demonstrated significant signal attenuation on an LTE mobile terminal. Specifically, the research group tested the 1.8 GHz band of GSM networks at Yuan-Ze University (YZU) by measuring the original radio power signals through mobile phones and comparing the transmission conditions during different weather circumstances.
\section{Dataset analytics}
In this section, we provide a brief overview of the data for the proposed study case and outline their key characteristics. The dataset consists of three distinct time series representing recorded signal levels under different weather conditions\footnote{The data is not publicly accessible as it is not under the ownership of the author. Interested parties are advised to reach out to the author, who will assist in making a formal request to access the data from the owner.}. Specifically, each time series is associated with a corresponding weather label: "sereno variabile" (i.e., changeable weather), "pioggia debole" (i.e., weak rain), and "pioggia moderata" (i.e., moderate rain). The first time series comprises 962 timesteps, while the last two time series consist of 137 and 122 timesteps, respectively. Since the number of images generated from the mapping process is slightly lower than the length of the input time series, the significant variance in the number of available timesteps poses a challenge to the dataset's practical utility. The primary challenge is not the imbalance in the dataset but rather the limited number of samples in two of the three classes for generating the image dataset.
\begin{figure}[htbp]
    \centering
    {\includegraphics[width=1\linewidth]{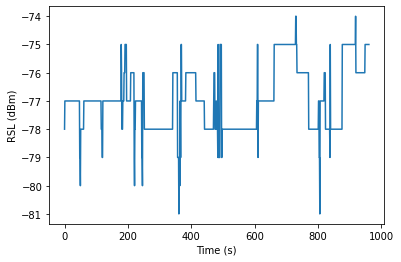}}
    \caption{Plot of received signal's level under changeable wheater conditions (i.e., "sereno variabile").}
    \label{fig:im1}
\end{figure}
\begin{figure}[htbp]
    \centering
    {\includegraphics[width=1\linewidth]{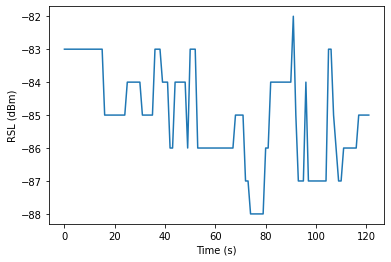}}
    \caption{Plot of received signal's level under moderate rain conditions (i.e., "pioggia moderata").}
    \label{fig:im2}
\end{figure}
\begin{figure}[htbp]
    \centering
    {\includegraphics[width=1\linewidth]{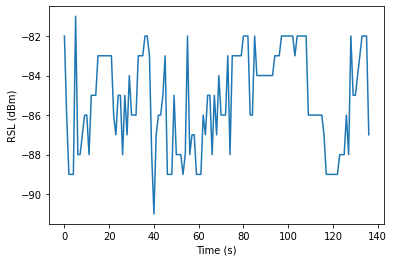}}
    \caption{Plot of received signal's level under weak rain conditions (i.e., "pioggia debole").}
    \label{fig:im3}
\end{figure}
\section{Data preprocessing}
\subsection{Mapping time series into images}\label{mapping}
The main novelty of the proposed approach is introducing a novel technique to encode time series into a series of images, capturing the temporal evolution of a fixed number of consecutive time steps. This transformation converts a time series $r_t \in R$ into a square image $I \in \mathcal{I}$ (where $I$ represents the set of images). To achieve this, a function $\sigma: R \mapsto \mathcal{I}$ transforms the time series into images, effectively visualizing the variable's evolution over time. Given a dataset $\mathcal{D}$ containing three time series $r_{t} \in R$, a new dataset of labeled images $\mathcal{D'}\,=\,\{\sigma(r_t)\,|\,\forall\,r_t \in \mathcal{D}\,\}$ is defined. Following this mapping procedure, the time series classification task is approached as an image classification task, utilizing convolutional neural networks.
\hfill \break \indent
A time series is a sequential record of observations for a specific variable, which, in this study, represents the measured level of received signals in decibel-milliwatts ($\si{\decibel}_m$). The original dataset $\mathcal{D}$ comprises temporal data, and it is clear that the temporal aspect contains vital features for addressing the time series classification problem. However, convolutional neural networks do not utilize this information as their filters operate in spatial dimensions and do not account for time. Additionally, data features must be densely represented, a condition not met unless the input space is low-dimensional.
\hfill \break \indent
The transformation function $\sigma: R \mapsto \mathcal{I}$, which converts time series to images, is implemented as follows. For a time series $(r_k )_t$, let $(\rho_k )_t$ represent the k-th variable at time \textit{t}, with $(\rho_k)_t = \frac{\partial (\rho_k)_t}{\partial t}$ and $(\theta_k)_t = \frac{\partial (\theta_k)_t}{\partial t}$ denoting the first and second time derivatives, respectively. The function aims to map variables with similar patterns close together and those with significantly different features far apart. This allows convolutional kernels to extract the required information to characterize the processed sample. Let us summarize the procedure in the following steps for clarity:
\begin{enumerate}[label=\Roman*)]
    \item \textit{Data normalization}: The initial step involves normalizing each time series within the $[0, n]$ range, where $n = 255$ represents the maximum value for pixel brightness in grayscale images.
    \item \textit{Numerical differentiation of the signal}: Since the data is discrete, we calculate the first and second derivatives using numerical differentiation. We conducted various tests to determine whether applying an \textit{i}-th point central difference, with $i = 3,5,7$, would enhance the model's performance in solving the classification task.
    \item \textit{Interpretation of the derivatives in polar coordinates}: We interpret the first derivatives as radii in a polar coordinate system with its origin at the image center. The values of $(\rho_k)_t$ are remapped to the interval $[0, I_e-1]$, where $I_e$ denotes the predetermined square dimensions of the image. Meanwhile, the absolute value of the second derivatives $|(\theta_k)_t|$ represents the angle in the polar coordinates system and is normalized to the range $[0, 2\pi]$.
    \item \textit{Image mapping procedure}
    Let $I_k$ represent a completely black grayscale image, meaning all the pixels have a value of 0. This can be easily depicted as an $N \times N$ square matrix, with each pixel identified by its position (row and column) and a value representing its brightness. To map a variable $(r_k)_t$ onto the image $I_k$, the value of each pixel denotes the intensity of the signal received at time $t$. The row $(i_k)_t$ and column $(j_k)_t$ of the image corresponding to the variable $(r_k)_t$ are computed as follows:
    \begin{equation}
        \begin{split}
        (i_{k})_{t} & = \floor*{(\rho_{k})_{t} \cdot cos(\theta_{k})_{t} + \frac{I_{e}}{2}}\\[10pt]
        (j_{k})_{t} & = \floor*{(\rho_{k})_{t} \cdot sin(\theta_{k})_{t} + \frac{I_{e}}{2}}
        \end{split}    
    \end{equation}
    \label{ComputePositions}
    To ensure clarity, we will demonstrate this step with a practical example. Let $t_k$ represent the time series that describes the level of the received signal $(r_k)_t$ over time. We aim to capture a portion of its temporal evolution in an image $I_k$. We map a specified number of time steps $l_w$ into each image to achieve this. Following the process outlined in the previous section, we define a temporal window $w_i$ with a fixed length of $l_w$ and map $l_w$ consecutive time steps into the image $I_k$. We then generate the next temporal window $w_{i+1}$ by excluding the first time step from the previous temporal window $w_i$ and including a new time step from the time series $t_k$. For instance, if $t_k$ is a time series consisting of ten timesteps (i.e., $t_k = t_1, t_2, t_3, t_4, t_5, t_6, t_7, t_8, t_9, t_{10}$), and the window length $l_w$ is set to 5, following the method above would result in mapping the time steps from $t_1$ to $t_5$ into the image $I_1$, the time steps from $t_2$ to $t_6$ into the image $I_2$, and so on. The process concludes when it is no longer feasible to include new time steps from the time series $t_k$. In this example, the last iteration would map the time steps from $t_5$ to $t_{10}$ into the image $I_5$, resulting in five images generated from the time series $t_k$. One of the reasons for using temporal windows is to strike a balance between the value of $l_w$ and the total number of images generated by the mapping process. Mapping too many time steps into a single image would produce a minimal dataset. We conducted multiple tests to determine the most suitable window length, as presented in section \ref{Experimental_phase}.
\end{enumerate}
\subsection{Limit cases: pixel superposition}
\label{PixelSuperpos}
\begin{figure}[htbp]
\subfloat[Sample from the class\newline"pioggiadebole".]{\label{fig:a}
\includegraphics[width=0.48\linewidth]{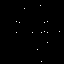}}
\subfloat[Sample from the class\newline"pioggiamoderata".]{\label{fig:b}
\includegraphics[width=0.48\linewidth]{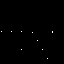}}\\
\subfloat[Sample from the class\newline "serenovariabile".]{\label{fig:c}
\includegraphics[width=0.48\linewidth]{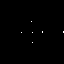}}
\subfloat[Extreme case of pixel\newline superposition.]{\label{fig:d}
\includegraphics[width=0.48\linewidth]{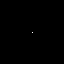}}%
\caption{Random image samples obtained from the \textit{time-series-to-images} procedure.}
\label{fig:quad1}
\end{figure}

Please note that the values of $(i_k)_t$ and $(j_k)_t$ are rounded down to the nearest integer, which may result in multiple variables being assigned to the same pixel (e.g., constant variables). This is referred to as "pixel superposition." The choice of image size $I_e$ affects the quantization error, as a smaller image size increases the likelihood of pixel superposition. As a design choice, we allow the conversion to overwrite values on the same pixels, since superposition occurs when the first derivative is very low (i.e., the features are not particularly relevant). Several experiments have shown that this mapping does not compromise the performance of the proposed method. In figure \ref{fig:quad1}, we present multiple image samples obtained by applying the aforementioned procedure to the provided case study.
\hfill \break \indent
\subsection{Formalization of the proposed approach}
The following lines provide a rigorous formalization of the proposed approach. Consider a general system $S$ whose internal state over time is characterized by a set of variables $V$, where $K=|V|$. In the presented case study, $K=1$, meaning that the image $I$ is comprised of a single channel filled with the normalized value of the received signal's level $(\hat{r}_k)_t=minmax((r_k )_t)$. To ensure a mathematically consistent formalism, the following definitions are provided:
\begin{equation}
\begin{split}
    \Lambda_t & = \{((i_k)_t, (j_k)_t)\;\;\;\;\forall k \in [0, K] \cap \mathbb{N}\} \\
    \overline{k}_{ijt} & \triangleq max\{k : (i_k)_t = i, (j_k)_t = j\} \\
    \chi_{ijt} & \triangleq (\hat{r}_{\overline{k}_{ijt}})_t \\
\end{split}
\end{equation}
From the aforementioned definitions, it follows that the scalar representing the timestep $t$ for each pixel at position $(i, j)$ and time is computed as:
\begin{equation*}
I_{ij}(t) \triangleq 
\left\{
\begin{alignedat}{2}
 \chi_{ijt}\;\;\;\; (i,j) \in \Lambda_t, \\
 0\;\;\;\; (i,j) \notin \Lambda_t, \\
\end{alignedat}
\right.
\end{equation*}
\indent
Please note that when $K > 1$, the formulation above can be expanded by representing the timestep $t$ as an RGB vector for each pixel at position $(i, j)$. To keep this explanation concise, we have not explicitly detailed this representation. Additionally, it is essential to note that specific values may fall outside the defined intervals. In such cases, these values are forced to the nearest allowed value.
\subsection{Balanced dataset generation}
\begin{figure}[htbp]
    \centering
    {\includegraphics[width=1\linewidth]{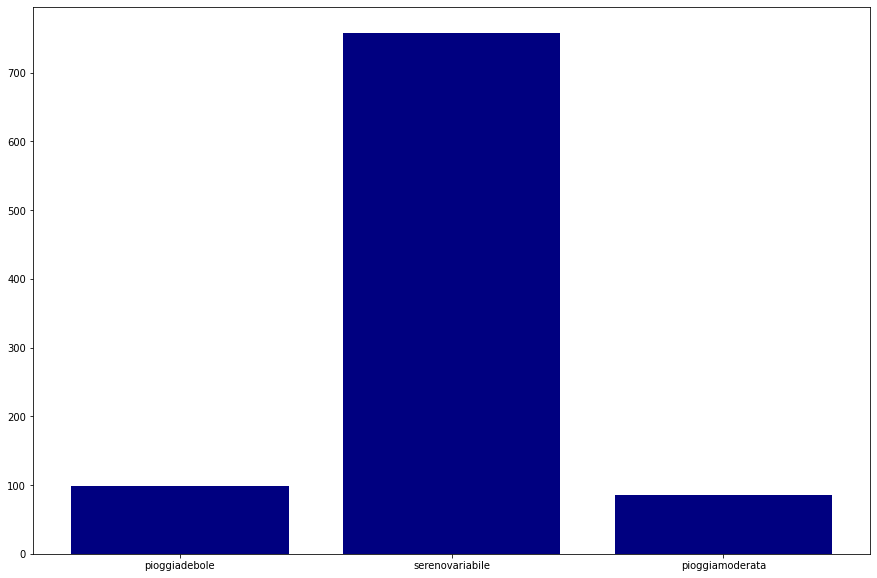}}
    \caption{Data distribution after applying the procedure without using data augmentation.}
    \label{fig:im4}
\end{figure}

The distribution of the dataset obtained after applying the procedure described in section \ref{mapping} is depicted in figure \ref{fig:im4}. The dataset is imbalanced, and consequently, it is difficult for the model to develop effective generalization capabilities. The model gets biased towards the majority classes in unbalanced dataset scenarios because they are overrepresented. Therefore, it becomes challenging for the model to correctly classify images belonging to the classes without enough training samples.
\newline
After applying the procedure with a definite window length, the number of samples available is about 125 images for the "weak rain" class, 110 for the "moderate rain" class, and 950 for the "changeable weather" class. The amount of generated images slightly decreases by increasing the window length parameter. Due to the intrinsic dataset imbalance, it was required to manipulate the dataset to obtain a new balanced image dataset that is finally fed to the model. We provide an automatic procedure to generate multiple datasets by including all sets of images from the first two classes and randomly selecting a set of images from the "changeable weather" to build a balanced dataset. Since the system allows data augmentation, the quantity of randomly extracted images is determined at runtime to match the number of samples from the other classes. Implementing this procedure was required to guarantee the execution of repeatable and unbiased tests.
\begin{figure}[htbp]
    \centering
    {\includegraphics[width=1\linewidth]{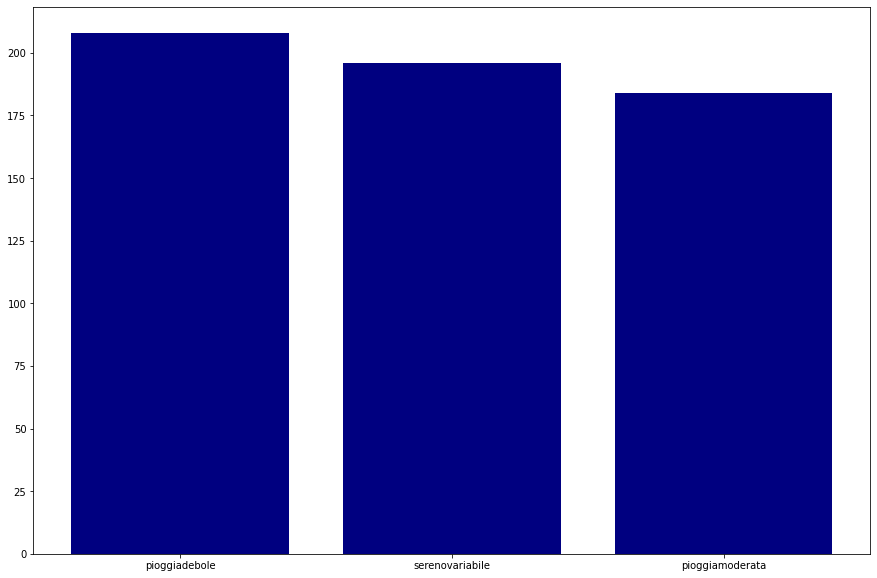}}
    \caption{Data distribution after applying the procedure using data augmentation to obtain a balanced dataset.}
    \label{fig:im5}
\end{figure}
\section{Model Architecture and Performance Evaluation}
The subsequent sections briefly examine the deep learning architecture utilized for the image classification task. Moreover, they outline the evaluation metrics chosen for assessing the model's performance and discuss the data augmentation techniques implemented to improve the model's robustness.
\subsubsection{Model Architecture and Performance Evaluation}
We leverage a conventional convolutional neural network (CNN) architecture for the image classification task. CNNs excel at extracting local spatial features from images and hierarchically combining them into higher-order representations, making them well-suited for this problem. Additionally, the inherent simplicity of the generated image data motivates the use of a simpler architecture. Complex models can be susceptible to overfitting on such data, potentially leading to performance degradation. Concerning the evaluation metrics, we employ categorical accuracy as an indicator to evaluate the model's convergence over time. However, evaluating image classification models on imbalanced datasets with limited samples necessitates metrics that capture performance beyond simple accuracy. While accuracy reflects the overall proportion of correct predictions, it can be misleading in scenarios where one class significantly outnumbers others. In such cases, a model could achieve high accuracy by trivially predicting the majority class for all instances. The F1 score offers a more nuanced evaluation by considering precision and recall. This balanced approach ensures the model performs well not only on the dominant class but also on the under-represented, yet potentially more critical, minority classes. 
\subsubsection{Data augmentation}
\label{DataAugment}
Deep neural networks often require large amounts of training data to provide reliable predictions, which are not always available. Therefore, a best practice is to augment the existing data to maximize the model's generalization capabilities. In particular, image augmentation in computer vision generally involves modifying the training images to generate a larger synthetic dataset to improve the model's downstream performance. The first data augmentation techniques explored during the project development are summarized in the following list:
\begin{itemize}
    \item Horizontal and vertical flipping (i.e., reversing the rows or columns of pixels in the case of a vertical or horizontal flip, respectively.)
    \item Height and width shifting (i.e.,  moving all image pixels in one direction, horizontally or vertically, while keeping the image dimensions the same).
    \item Rotation, Zooming, brightness range clipping (i.e., rotating, zooming the image, and modifying its brightness range).
    \item Shearing (i.e., a linear map that displaces every point in a fixed direction by an amount proportional to its signed distance from the parallel line to that direction and passes through the origin). 
    \item Random crop \cite{zhong2020random} (i.e., creating a random subset of the original image). As the authors point out, this technique enables the model to learn the features of objects that are not wholly visible in the image or present the same scaling of the training data.
    \item Random erasing \cite{takahashi2019data} (i.e., randomly selecting a rectangle region in an image and erasing its pixels with random values). This method generates training images with different occlusion levels, which decreases the risk of overfitting and makes the model robust to occlusion.
\end{itemize}
However, data augmentation procedures should be chosen carefully depending on the dataset. For instance, horizontal and vertical image flipping must be discarded in solving a task such as a chest X-ray image classification problem for early COVID-19 detection \cite{ghaleb2021covid}. Horizontally flipping the image would result in positioning the patient's heart to indicate a rare congenital disorder known as \textit{situs inversus}. On the contrary, changing an image's brightness range or contrast for an object classification task would not alter its feature since each image could be taken from multiple angles and light conditions. \label{QuandoUsare} \newline \indent
Figure \ref{fig:im4} illustrates the distribution of the dataset following the application of data augmentation. It is clear that the data augmentation process has led to a balanced dataset, thus enhancing the model's capacity to develop effective generalization capabilities.

\section{Experimentation phase}
\label{Experimental_phase}
The following section presents and discusses the outcomes of the experiments conducted. The analysis aims to identify the most influential factors affecting the model's performance from the various factors explored. We have conducted multiple experiments to determine the most effective combinations of data augmentation techniques and other elements. As outlined in the previous section (see \ref{DataAugment}), most of the examined data augmentation techniques did not yield substantial improvements and, in some cases, degraded the model's performance. For brevity, we only present the results of the combinations that have shown significant outcomes. In conclusion, we propose the configuration that has been determined as the optimal choice in terms of performance and simplicity.
\subsection{Window length} 
The window length parameter is critical in determining the number of timesteps mapped into a single image. It also plays a crucial role in the effectiveness of the proposed approach, as revealed by experiments. The rationale behind this research direction is as follows: the window length dictates the quantity of information mapped in each image. Consequently, an increase in the number of time steps mapped into a single image makes it easier for the model to extract information about the temporal evolution of the time series. For example, when mapping a few time steps, each image contains only short-term information. To illustrate, if the input time series has ten timesteps and the window length is set to five timesteps, the first image represents the relationships between timesteps 1-5, the second image describes the relationships between timesteps 2-6, and so forth. Therefore, determining an appropriate window length enables the model to capture long-term dependencies.
Additionally, when the number of mapped timesteps is limited, the resulting features are sparse, especially if the image edge is not chosen correctly. Traditional convolutional neural networks struggle when processing images with sparse features in this context. To address this issue, we have chosen to employ longer temporal windows to create images with dense features to the fullest extent the proposed method allows. Due to pixel superposition, as described in section \ref{PixelSuperpos}, it has been crucial to establish an appropriate image edge that can balance the number of samples within the image with its dimensions. Mapping an excessive number of samples into an image with minimal edges would significantly increase the occurrence of pixel superpositions.
\label{Window_length}
\subsection{Random erasing}
\label{RandomErasingExp}
Random erasing is an innovative data augmentation technique introduced by Zhong et al. \cite{zhong2020random} for training convolutional neural networks. This technique involves randomly selecting a rectangle region within an image during the training phase and replacing its pixels with random values. The process generates training images with varying levels of occlusion, which reduces the risk of overfitting. Random erasing complements commonly used data augmentation techniques such as random cropping and flipping. It is widely applied in computer vision to improve the generalization capabilities of vision systems, particularly in making them robust to occlusion, which is common in various applications, including robotics. Although the intuition behind utilizing this technique is not directly related to occlusion, it shares similarities with the application of Generative Adversarial Networks. The concept behind GANs involves training two models, the generator and discriminator, simultaneously. The generator produces a batch of samples that are presented to the discriminator to be classified as real or fake, alongside real examples from the domain. The discriminator is then updated to better distinguish real and fake samples in the next iteration, while the generator is updated based on how well the generated samples deceive the discriminator. This concept has inspired the application of random erasing to the CNN in the proposed method. In simple terms, we apply a patch to a portion of the image to make the model learn to discriminate whether an image belongs to a particular class, even without having complete information regarding the entire distribution of the image. Additionally, this method has a positive side effect: it encourages the model to complete the classification task with limited computational power and reduces the computations needed by the CNNs to process images. However, in practical terms, the reduction in computations due to this aspect is marginal. It's important to note that incorporating this method into the proposed approach requires a slight modification of the random erasing procedure. The traditional use of random erasing involves adding patches to the images whose color matches the portion of the image that is being covered. However, in this case, this approach would be counterproductive, as it would introduce large patches of pixels that the model could easily interpret as features. To address this, we have adapted the technique to only insert black patches. This adjustment was made because a pixel that has not been assigned a timestep is completely black, so using black patches allows us to cover portions of the image without introducing information that the model could mislead the model.
\subsection{Results discussion}
The following section presents the results, which are summarized in table \ref{results_table}. The importance of window length and the advantages of random erasing are highlighted through a detailed analysis of the experimental results shown in Table 1. Initially, the results achieved with minimal window lengths are presented to illustrate the model's initial performance. This is followed by a demonstration of the impact of gradually increasing the window lengths on the model's performance. Furthermore, the outcomes obtained with and without data augmentation techniques for each window length are emphasized, providing a comprehensive understanding of the benefits of combining different data augmentation techniques at various levels, particularly on the time series and the resulting image data. The subsequent discussion details our results, elaborating on the observed performance trends across different experimental setups. 
Initial findings from experiments 1 and 2 revealed that representing a small number of timesteps in each image resulted in unsatisfactory performance. This was an expected outcome, as the limited temporal context hindered the model's ability to capture underlying patterns. However, implementing data augmentation at various levels significantly enhanced the model's performance under these conditions. Building on these results, experiments 3 and 4 validated that increasing the number of timesteps represented in each image improved the model's predictive capabilities on the test set. Data augmentation proved highly effective in this scenario, significantly improving the model's accuracy. These findings indicate that the additional temporal information, combined with improved generalization from data augmentation, is essential for the model's success. Experiments 5 and 6 delved into the impact of window length on the model's performance. It was found that increasing the window length led to quicker and more stable convergence, although it did not result in substantial performance improvements. Experiment 5 exhibited overfitting; however, the implementation of data augmentation in Experiment 6 mitigated this issue, emphasizing the importance of data augmentation in preventing overfitting. Experiments 7 and 8 determined that mapping at least 30 timesteps into each image is crucial for the model to show satisfying performance. Experiments 9 and 10 indicated that mapping 40 timesteps into each image is adequate to achieve around 90\% accuracy in the classification task. Experiments 11 and 12 highlighted the negative impact of the pixel superposition phenomenon. The findings demonstrated that mapping numerous timesteps (e.g., 50) into an image with dimensions of only 64x64 results in a performance bottleneck due to information loss. Consistent with previous observations, data augmentation enhanced the model's generalization capabilities. To comprehensively understand the results, we compared them with an experiment involving doubled image dimensions. In this final experiment, the performance of the best setup identified from the previous experiments was evaluated. Increasing the image size alleviated the adverse effects of pixel superposition, ultimately boosting the model's performance and enabling it to achieve 95\% accuracy on the test set. Further tests were carried out with larger image dimensions, and it was found that when not enough timesteps were mapped in extremely large images without significant performance improvement that could be attributed to the sparse features in the images. 
\begin{table*}[t]
\label{tab:results}
\centering
\resizebox{\textwidth}{!}{%
\begin{tabular}{|c|c|c|c|c|c|}
\hline
\textbf{Experiment} & \textbf{Window Length} & \textbf{Data Augmentation} & \textbf{Image Edge} & \textbf{Accuracy Test} & \textbf{F1-score} \\ \hline
1 & 5 & - & 64 & 0.56 & 0.53 \\ \hline
2 & 5 & MA, RE & 64 & 0.61 & 0.52 \\ \hline
3 & 10 & - & 64 & 0.56 & 0.53 \\ \hline
4 & 10 & MA, RE & 64 & 0.70 & 0.71 \\ \hline
5 & 15 & - & 64 & 0.72 & 0.74 \\ \hline
6 & 15 & MA, RE & 64 & 0.72 & 0.73 \\ \hline
7 & 30 & - & 64 & 0.69 & 0.77 \\ \hline
8 & 30 & MA, RE & 64 & 0.81 & 0.86 \\ \hline
9 & 40 & - & 64 & 0.89 & 0.916 \\ \hline
10 & 40 & MA, RE & 64 & 0.90 & 0.92 \\ \hline
11 & 50 & - & 64 & 0.90 & 0.90 \\ \hline
12 & 50 & MA, RE & 64 & 0.91 & 0.92 \\ \hline
13 & 50 & - & 128 & 0.93 & 0.93 \\ \hline
14 & 50 & MA, RE & 128 & \textbf{0.95} & \textbf{0.95} \\ \hline
\end{tabular}%
}
\caption{Experimental results: impact of window length and data augmentation on model performance. 
\label{results_table}
\newline Legend: MA - Moving average (window length = 3), RE - Random erasing}
\end{table*}
\begin{figure}[htbp]
    \centering
    {\includegraphics[width=1\linewidth]{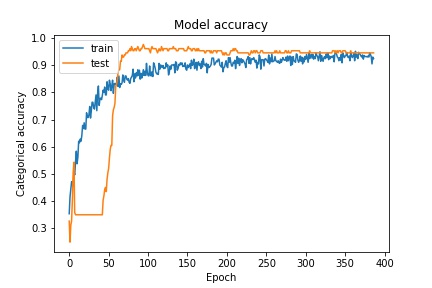}}
    \caption{Categorical accuracy reported by the best setup (experiment 14).}
\end{figure}
\begin{figure}[htbp]
    \centering
    {\includegraphics[width=1\linewidth]{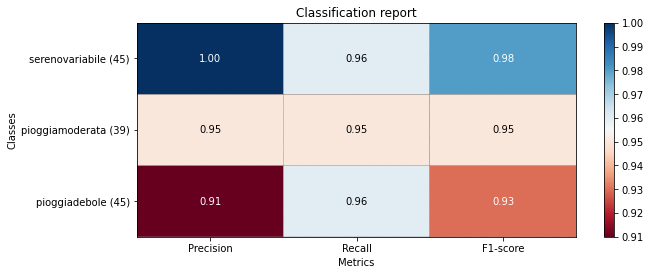}}
    \caption{Precision, recall, and F1-score reported by the best setup (experiment 14).}
\end{figure}
\section{Conclusions}
In this paper, we have investigated the problem of identifying specific weather phenomena from the received signal level (RSL) in 4G/LTE mobile terminals. By utilizing time-series data representing RSL, we have introduced a novel method to encode time series as images. We have framed the task as an image classification problem, effectively addressing it using convolutional neural networks (CNNs). Initially, we have outlined the issue of rainfall estimation and its significance in real-world applications, particularly its impact on electromagnetic waves and cellular network performance. Subsequently, we have comprehensively analyzed the dataset used in this study. We have then elaborated on transforming time series into images, rigorously formalizing the approach. During the experimental phase, we concentrated on the image classification task, demonstrating the advantages of employing CNNs. Furthermore, we have examined the most influential factors affecting the model's performance, such as window length and data augmentation techniques. The experiments have confirmed the validity of our approach, showcasing the model's capability to achieve high accuracy in the classification task.
Specifically, we showcased the benefits of combining various data augmentation techniques on the time series and the generated image data at different levels. Our study presented the advantages of integrating various data augmentation methods for both time series and generated image data at different stages. The outcomes emphasized the substantial enhancements in model performance achieved through these techniques, underscoring their pivotal role in developing resilient models. We aim for this study to spark interest within the research community to explore diverse techniques for transforming time series into images. In future works, we plan to compare the proposed approach with other established methods for encoding time series data into images, such as Gramian Angular Fields \cite{wang2015imaging} and Markov Transition Fields \cite{wang2015encoding}. Furthermore, we intend to apply this method to other datasets and more complex time series classification tasks in future research.
\newpage
\bibliographystyle{unsrt}  
\bibliography{references}
\end{document}